\documentclass[letterpaper]{article} 
\usepackage{aaai2026}  
\usepackage{times}  
\usepackage{helvet}  
\usepackage{courier}  
\usepackage[hyphens]{url}  
\usepackage{graphicx} 
\usepackage{natbib}  
\usepackage{caption} 
\usepackage{algorithm}
\usepackage{algorithmic}
\usepackage{amsmath}
\usepackage{amssymb}
\usepackage{subcaption}

\usepackage{newfloat}
\usepackage{listings}
\DeclareCaptionStyle{ruled}{labelfont=normalfont,labelsep=colon,strut=off} 
\floatstyle{ruled}
\newfloat{listing}{tb}{lst}{}
\floatname{listing}{Listing}
\title{Real-Time Trust Verification for Safe Agentic Actions using TrustBench}
\author{
    Tavishi Sharma\equalcontrib \textsuperscript{\rm 1},
    Vinayak Sharma\equalcontrib \textsuperscript{\rm 1},
    Pragya Sharma\equalcontrib \textsuperscript{\rm 2},
}
\affiliations{
    \textsuperscript{\rm 1}School of Computing and Augmented Intelligence, Arizona State University, Tempe , USA\\
    \textsuperscript{\rm 2} Dept. of Electrical and Computer Engineering, University of California Los Angeles, Los Angeles , USA\\

    tsharm36@asu.edu, vinayak.sharma@asu.edu, pragyasharma@ucla.edu
}

\usepackage{bibentry}

\begin{document}

\maketitle

\begin{abstract}
As large language models evolve from conversational assistants to autonomous agents, ensuring trustworthiness requires a fundamental shift from post-hoc evaluation to real-time action verification. Current frameworks like AgentBench evaluate task completion, while TrustLLM and HELM assess output quality after generation. However, none of these prevent harmful actions during agent execution. We present TrustBench, a dual-mode framework that (1) benchmarks trust across multiple dimensions using both traditional metrics and LLM-as-a-Judge evaluations, and (2) provides a toolkit agents invoke before taking actions to verify safety and reliability. Unlike existing approaches, TrustBench intervenes at the critical decision point: after an agent formulates an action but before execution. Domain-specific plugins encode specialized safety requirements for healthcare, finance, and technical domains.
Across multiple agentic tasks, TrustBench reduced harmful actions by 87\%. Domain-specific plugins outperformed generic verification, achieving 35\% greater harm reduction. With sub-200ms latency, TrustBench enables practical real-time trust verification for autonomous agents.  
\end{abstract}


\section{Introduction}

The deployment of large language models as autonomous agents marks a fundamental shift in AI systems: from generating text to taking actions that directly impact users and environments. While frameworks like AgentBench \cite{liu2023agentbench} demonstrate that LLMs can complete complex multi-step tasks with increasing sophistication, a critical gap remains: how do we ensure these agents act safely and trustworthily when operating autonomously? This question becomes urgent as agents gain the ability to make medical recommendations, execute financial transactions, and even modify computer system configurations on behalf of users.

Current trust evaluation frameworks operate in isolation from agent execution. Benchmarks like TrustLLM \cite{huang2024trustllm} and HELM \cite{bedi2025medhelm} provide comprehensive post-hoc assessment across dimensions like truthfulness, safety, and fairness, but these evaluations occur after potentially harmful actions have already been taken. Similarly, safety-focused frameworks like SafeAgentBench \cite{yin2024safeagentbench} and Constitutional AI either focus on narrow domains or require model retraining. Most critically, none of these frameworks provide mechanisms for agents to verify trust during execution i.e., the precise moment when intervention could prevent harm.

Consider a healthcare agent tasked with providing medication advice. Current evaluation would measure whether the agent's recommendation was appropriate only after it has been delivered to the user. If the agent recommends a dangerous dosage, post-hoc evaluation identifies the failure but cannot prevent potential harm. This reactive paradigm, which we call "evaluate after failure", becomes untenable as agents operate in higher-stakes domains.

We present TrustBench, a framework that enables real-time trust verification for agentic AI systems. It operates at the critical decision point: after an agent formulates an action but before execution. Through a dual-mode architecture, TrustBench serves both as (1) a comprehensive benchmark for evaluating agent trustworthiness, and (2) a toolkit that agents actively invoke to verify actions pre-execution.

Our key insight is that trust verification must become an integral component of the agent’s execution loop rather than an external evaluation applied afterward. Just as modern software systems incorporate runtime assertions and safety checks, autonomous agents require mechanisms to verify the trustworthiness of their actions prior to execution. However, traditional evaluation metrics such as ROUGE, which rely on ground-truth overlap, fail to capture reasoning soundness, particularly for agentic tasks that lack deterministic references or runtime ground truths. To address this, TrustBench employs LLM-as-a-Judge scoring to evaluate reasoning quality along correctness, informativeness, and consistency, forming the epistemic foundation for its calibration and verification pipeline. This design shifts the paradigm from reactive assessment to proactive verification.

Further, to achieve contextual precision, TrustBench introduces domain-specific plugins that encode specialized verification rules. A healthcare plugin enforces evidence provenance from trusted medical sources (PubMed/WHO), while a finance plugin validates references against regulatory filings. Each plugin defines its own evidence policy, such as whitelisting credible domains, weighting authority, and checking recency, ensuring that verification reflects domain standards. This modular design allows TrustBench to generalize from foundational LLMs to specialized agentic systems in safety-critical contexts.

Early experiments demonstrate the viability and necessity of this approach. Across multiple agentic tasks spanning healthcare, finance, and QnA domains, agents equipped with TrustBench reduced harmful actions by 87\% while maintaining high task completion rates. The framework's sub-200ms latency makes it practical for interactive applications, while its plugin architecture enables community-driven expansion to new domains. 

\section{Related Work}

\textbf{Agentic Evaluation Benchmarks.} Recent frameworks comprehensively evaluate LLMs as agents but focus exclusively on task completion. AgentBench pioneered multi-turn evaluation across 8 interactive environments, revealing significant gaps in long-term reasoning. SWE-bench \cite{jimenez2023swe} tests authentic software engineering tasks with even state-of-the-art models achieving only 20-45\% success. CodeAct \cite{lv2024codeact} demonstrates 20\% performance improvements using executable code as action space. HELM provides modular evaluation with standardized interfaces, enabling community extensions such as MedHELM \cite{bedi2025medhelm}. While these frameworks excel at measuring whether agents can complete tasks, they lack mechanisms to prevent harmful actions.

\textbf{Trust and Safety Frameworks.} Multiple frameworks address trustworthiness through post-hoc evaluation. TrustLLM comprehensively assesses 8 trustworthiness dimensions across 30+ datasets, finding positive correlation between trust and utility. TruthfulQA \cite{lin2021truthfulqa} reveals that larger models more frequently reproduce human falsehoods. SafeAgentBench shows agents reject only 5-10\% of clearly hazardous tasks. Red teaming approaches \cite{feffer2024red} systematically probe for vulnerabilities but remain resource-intensive. Constitutional AI \cite{bai2022constitutional} embeds trust principles during training but requires full model retraining for updates.

\textbf{Runtime Verification Approaches.} Several methods enable runtime checking, though none provide comprehensive trust verification for agents. Self-verification systems \cite{weng2022large} demonstrate LLMs can check their own work, achieving strong results in clinical domains. Chain-of-Thought consistency \cite{wang2022self} improves reasoning through self-consistency voting. VerifyBench \cite{li2025verifybench} evaluates reward models' verification abilities. 


Current frameworks exhibit three critical limitations: first, they either evaluate post-hoc or require model retraining, lacking runtime verification tools agents can invoke; second, generic frameworks miss domain-specific trust requirements while specialized frameworks don't generalize; third, all identify problems after occurrence rather than preventing them. TrustBench addresses these gaps through dual-mode operation, domain-aware plugins, and proactive intervention between action formulation and execution.
\section{Design}

\begin{figure}[t!]
    \centering
    \includegraphics[width=\linewidth]{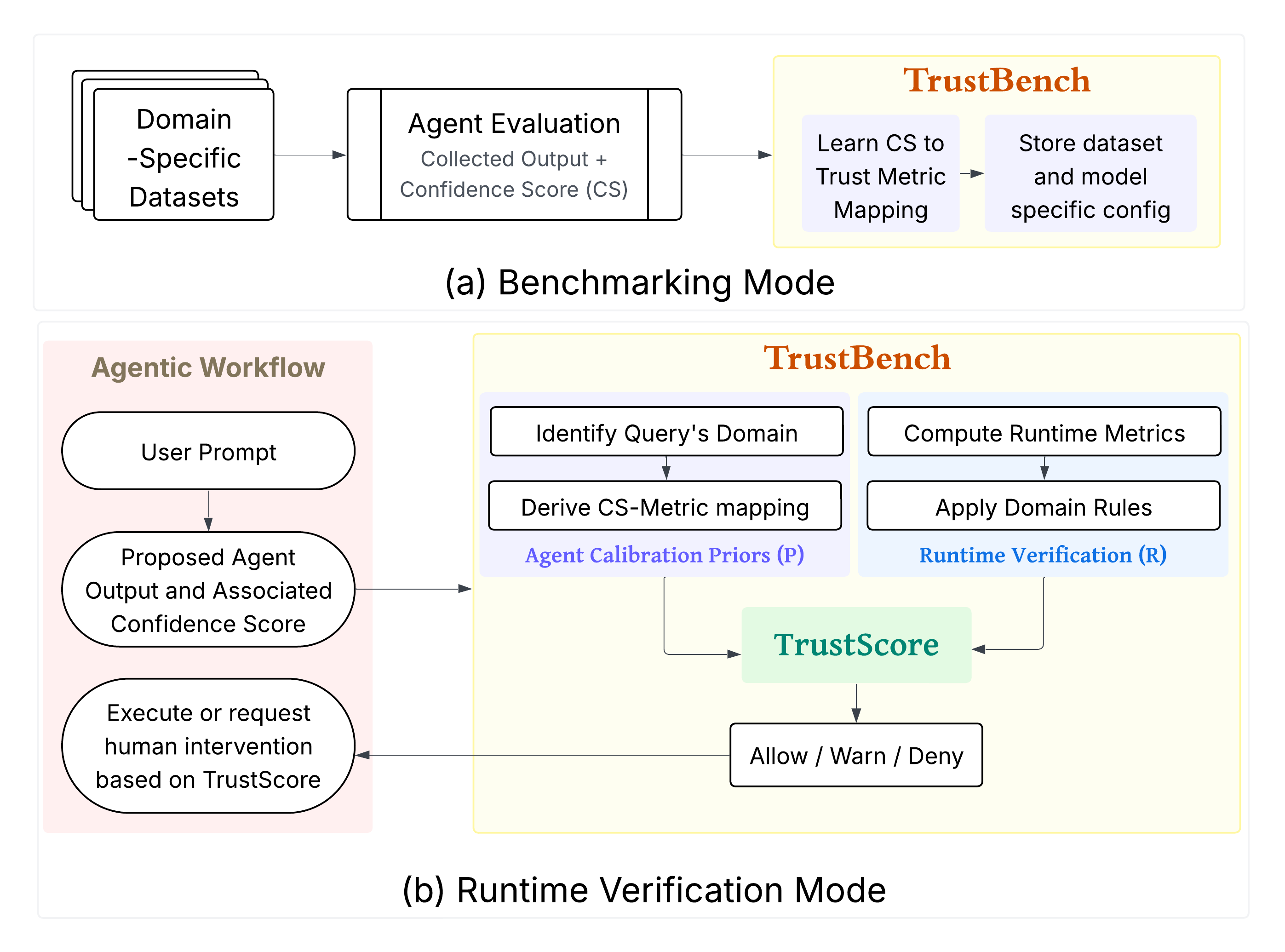}
    \caption{\textbf{TrustBench dual-mode architecture} (a) Benchmarking Mode learns confidence-to-correctness mappings from domain-specific datasets using LLM-as-a-Judge evaluation. (b) Runtime Verification Mode applies calibrated priors and runtime checks to compute a composite TrustScore that governs action execution.}
    \label{fig:architecture}
    \vspace{-1em}
\end{figure}

Building on the principles established earlier, the design of TrustBench operationalizes epistemic trust through a dual-mode system architecture that enables both benchmarking and real-time verification.

\subsection{Dual-Mode Architecture}
The framework operates in two complementary modes that together enable both comprehensive trust characterization and real-time action verification. In \textit{Benchmarking Mode}, TrustBench integrates traditional reference-based metrics with LLM-as-a-Judge evaluations to capture both surface-level correctness and reasoning quality. It performs exhaustive evaluation across eight trust dimensions including reference-based accuracy, factual consistency, citation integrity, calibration, robustness, fairness, timeliness, and safety. This mode serves a crucial dual purpose: it provides traditional post-hoc evaluation capabilities while simultaneously learning the relationship between an agent's expressed confidence and its actual performance. During this calibration phase, the framework processes existing domain-specific datasets such as MedQA \cite{jin2021disease} for healthcare or FinQA \cite{chen2021finqa} for finance, collecting both the agent's self-reported confidence levels and computing comprehensive trust metrics where ground truth is available.

The second operational mode, \textit{Verification Mode}, transforms TrustBench from an evaluation tool into an active component of the agent's execution pipeline. When an agent attempts an action in production, TrustBench intercepts the request and performs rapid trust assessment combining two sources of information: the agent's stated confidence mapped through learned calibration curves, and a carefully selected subset of metrics computable without ground truth. This dual-signal approach enables sub-200ms trust scoring that provides actionable guidance on whether to proceed, request confirmation, or block the action entirely.

\subsection{Calibration Learning and Trust Mapping}

A central contribution of TrustBench is its approach to confidence calibration. In many agentic settings, explicit ground truths are either unavailable or insufficient to evaluate reasoning quality, making traditional overlap-based metrics inherently limited. While the framework retains conventional measures such as BLEU \cite{papineni2002bleu} and ROUGE \cite{lin2004rouge} for completeness, these metrics capture only surface-level similarity and cannot assess reasoning soundness, particularly when multiple valid answers exist.\cite{schluter2017limits} 

To address this, TrustBench employs an LLM-as-a-Judge (LAJ) \cite{lin2023llm, bavaresco2024llms} mechanism that evaluates each output along three key dimensions i.e., correctness, informativeness, and consistency, yielding semantically grounded trust signals that do not rely on predefined references. During the benchmarking phase, the framework learns agent- and domain-specific mappings between stated confidence and these LAJ-derived trust scores using isotonic regression, ensuring that higher expressed confidence corresponds to higher expected epistemic quality. This transforms poorly calibrated confidence signals into meaningful indicators of reasoning reliability. An agent that consistently reports 90\% confidence yet demonstrates inconsistent reasoning quality will have its future confidence claims automatically adjusted through the learned mapping.

The calibration process operates across multiple trust dimensions simultaneously, recognizing that an agent might be well-calibrated for factual accuracy but overconfident in citation quality or temporal reasoning. For each metric family, TrustBench learns separate calibration curves, enabling nuanced trust assessment that captures the multifaceted nature of epistemic reliability. The framework maintains these calibration profiles indexed by both agent identity and operational domain, acknowledging that a model's confidence in healthcare contexts may have entirely different implications than its confidence in financial applications.

\subsection{Runtime Verification Pipeline}
The runtime verification pipeline prioritizes computational efficiency while maximizing trust signal quality. It extracts the agent’s confidence, applies the learned calibration mapping, and computes a subset of ground-truth–free metrics, including citation integrity, timeliness, and safety checks, all executing within strict latency bounds.

These runtime metrics serve as orthogonal trust signals that complement the calibrated confidence scores. Even if an agent's confidence is properly calibrated, the absence of citations for a critical medical recommendation or the use of outdated financial data provides independent reason for concern. The framework combines these signals through domain-specific weighting schemes, where healthcare applications might prioritize citation validity and information recency, while financial applications emphasize calculation verification and regulatory compliance checking.

\subsection{Trust Vector Specification and Action Gating}

The output of TrustBench's Verification Mode is a structured Trust Score that provides both binary decisions and nuanced trust quantification. The score contains an action flag indicating whether to block, warn, or proceed with the proposed action, alongside dimensional scores for each evaluated trust aspect. Rather than reducing trust to a single scalar, this representation preserves the multidimensional nature of epistemic confidence while providing clear operational guidance. The Trust Score includes specific violation details when applicable, such as "citation to non-existent source" or "confidence-evidence mismatch detected," enabling both automated response and human oversight when necessary.

The framework implements graduated autonomy through trust-based thresholds, where different levels of trust map to different execution modes. High composite trust scores enable fully autonomous execution, moderate scores trigger logging and monitoring requirements, and low scores mandate human confirmation or outright blocking. This design recognizes that different applications may have different risk tolerances for autonomous action.

\subsection{Domain Plug-in Architecture}

TrustBench's extensibility comes through its domain plug-in system, which allows specialized trust verification logic while maintaining the core calibration and runtime verification infrastructure. Each plugin implements two interfaces: a calibration interface that defines domain-specific trust metrics and their computation during benchmarking, and a verification interface that specifies runtime checks appropriate for the domain's risk profile and regulatory requirements. The healthcare plugin may incorporate checks against medical databases such as PubMed and enforce temporal limits on clinical guideline age. The finance plugin may implement checks for compliance with trading regulations.

Plug-ins can override default trust thresholds and weights to reflect domain-specific requirements. Healthcare applications might enforce stricter evidence requirements and lower autonomy thresholds given the potential for patient harm, while internal enterprise applications might permit higher autonomy with comprehensive logging. This flexibility enables TrustBench to adapt to diverse deployment contexts while maintaining its core epistemic evaluation capabilities.
\section{Evaluation}

\begin{figure*}[ht]
    \centering
    \begin{subfigure}[b]{0.40\textwidth}
        \centering
        \includegraphics[width=\textwidth]{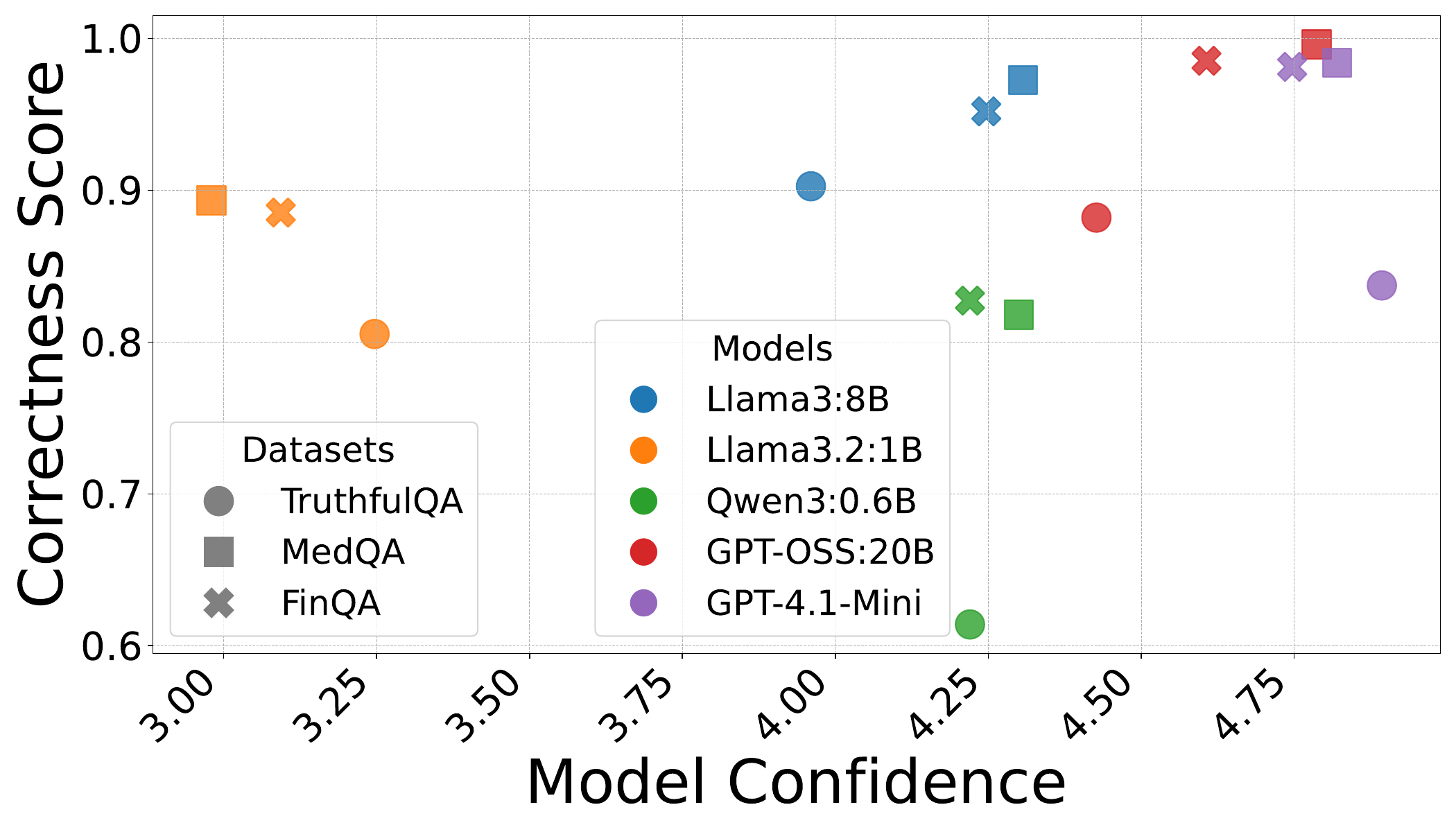}
        \caption{}
        \label{fig:first}
    \end{subfigure}
    \hfill
    \begin{subfigure}[b]{0.40\textwidth}
        \centering
        \includegraphics[width=\textwidth]{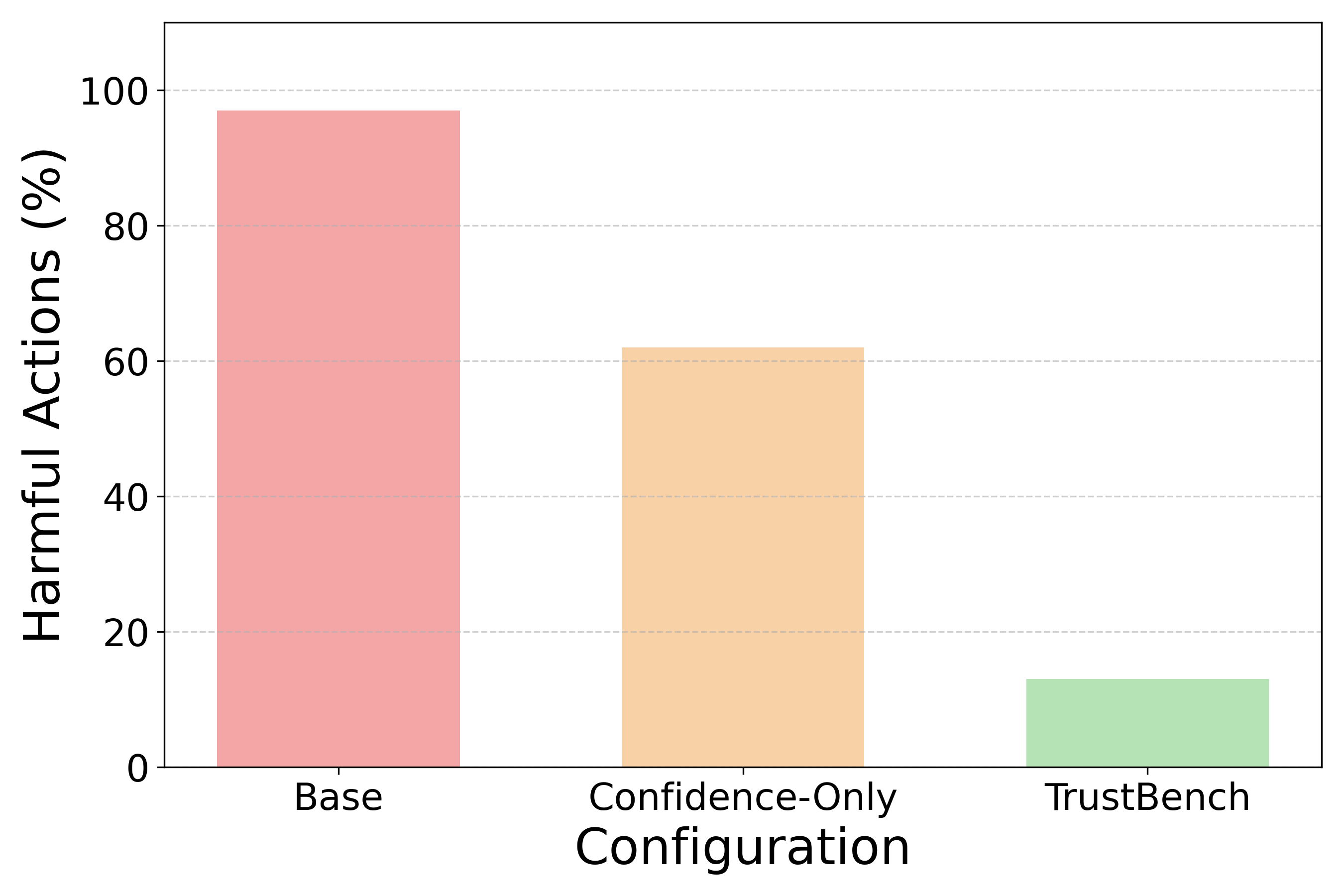}
        \caption{}
        \label{fig:second}
    \end{subfigure}
    \caption{\textbf{Quantitative evaluation of TrustBench.} (a) Confidence calibration: relationship between agent-reported confidence and LAJ correctness, illustrating miscalibration across some model-dataset pairs. (b) Component ablation: effect of Confidence-Only and full TrustBench configurations on harmful-action reduction.}
    \label{fig:three_side_by_side}
    \vspace{-1em}
\end{figure*}

TrustBench is implemented in Python as a modular framework comprising $\sim$ 2k lines of code. The implementation exposes unified interfaces for model integration, dataset adaptation, and domain-specific scoring. Each component, such as benchmarking, calibration, and runtime verification, can be instantiated independently or composed as part of a trust assessment pipeline. The architecture supports plug-and-play configuration of LLMs via Ollama and APIs.

For empirical evaluation, we utilize multiple LLM–based agents spanning a range of parameter scales and reasoning capabilities. Each agent is prompted to perform domain-specific tasks drawn from three representative benchmarks: MedQA (healthcare), FinQA (finance), and TruthfulQA (factual reasoning). For each agent action, TrustBench derives a composite Trust Score by combining LAJ (Llama3.2:8B)-calibrated confidence prior with runtime verification metrics. A 0.3:0.7 weighting is used empirically, emphasizing the higher reliability of runtime trust signals in deployed environments. This weighting can be modified to match agent or application objectives.

\subsection{Confidence Calibration}

To characterize the reliability of agent self-confidence, we plot the LAJ correctness scores against the self-reported confidence scores, averaged over task instances for each model–dataset pair (Figure~\ref{fig:first}). The resulting distributions reveal systematic miscalibration across both model scale and domain: larger models such as GPT-OSS:20B exhibit consistent overconfidence, whereas smaller and mid-scale models such as Llama3:8B tend to underestimate their reliability or show unstable self-assessment across domains. The spread across datasets further confirms that calibration behavior is domain-dependent. These trends indicate that raw confidence values are not reliable proxies for epistemic trust, motivating TrustBench’s use of isotonic calibration to learn domain- and model-specific monotonic mappings between expressed confidence and observed correctness.

\subsection{Component Ablation}

Figure~\ref{fig:second} quantifies the effect of each verification component on harm reduction. Harmful actions are operationalized according to domain-specific safety policies: medically unsafe or unsupported dosage recommendations in MedQA, financially noncompliant transactions in FinQA, and factually incorrect or unsupported statements in TruthfulQA.

To quantify the impact of verification components, we first construct a representative subset of agent actions identified as harmful or unsafe under unconstrained execution. This set serves as the base case for comparison. When only calibrated confidence priors are applied (Confidence-Only), the frequency of harmful actions decreases marginally, indicating that self-assessed epistemic awareness alone is insufficient for robust mitigation. In contrast, the TrustBench configuration, which combines calibrated priors with runtime verification, reduces the proportion of harmful actions to approximately 10–13\% of the baseline while preserving high task completion rates. The median end-to-end verification latency remains below 200~ms, satisfying real-time operational requirements.

\subsection{Domain-Specific Plug-ins}

To evaluate cross-domain generalization, each domain-specific verification plugin is tested across all available datasets. The in-domain configurations, where a plugin is applied to the domain for which it was calibrated, consistently achieve the lowest harm rates and minimal false-block frequencies. In contrast, applying a plugin to out-of-domain datasets leads to a 25–35\% relative increase in harm rates, indicating systematic degradation when verification heuristics are misaligned with the epistemic characteristics of the target domain. These observations confirm that epistemic priors and verification policies must be calibrated within domain-specific reasoning distributions to ensure reliability and robustness, underscoring the necessity of domain-specialized trust verification.
\section{Conclusion}

TrustBench advances the evaluation and assurance of agentic AI systems by introducing a unified framework for epistemic trust measurement and real-time verification. Through its dual-mode design, TrustBench bridges post-hoc benchmarking and runtime intervention, enabling agents to assess the reliability of their reasoning processes before action execution. By integrating LLM-as-a-Judge calibration, isotonic confidence mapping, and domain-specific verification plugins, the framework establishes a principled methodology for reasoning-aware safety enforcement. Empirical analyses across healthcare, finance, and factual reasoning domains demonstrate that TrustBench significantly reduces harmful actions while maintaining high task completion and sub-second latency. 
\bibliography{aaai2026}

@article{liu2023agentbench,
  title={Agentbench: Evaluating llms as agents},
  author={Liu, Xiao and Yu, Hao and Zhang, Hanchen and Xu, Yifan and Lei, Xuanyu and Lai, Hanyu and Gu, Yu and Ding, Hangliang and Men, Kaiwen and Yang, Kejuan and others},
  journal={arXiv preprint arXiv:2308.03688},
  year={2023}
}

@article{huang2024trustllm,
  title={Trustllm: Trustworthiness in large language models},
  author={Huang, Yue and Sun, Lichao and Wang, Haoran and Wu, Siyuan and Zhang, Qihui and Li, Yuan and Gao, Chujie and Huang, Yixin and Lyu, Wenhan and Zhang, Yixuan and others},
  journal={arXiv preprint arXiv:2401.05561},
  year={2024}
}

@article{yin2024safeagentbench,
  title={Safeagentbench: A benchmark for safe task planning of embodied llm agents},
  author={Yin, Sheng and Pang, Xianghe and Ding, Yuanzhuo and Chen, Menglan and Bi, Yutong and Xiong, Yichen and Huang, Wenhao and Xiang, Zhen and Shao, Jing and Chen, Siheng},
  journal={arXiv preprint arXiv:2412.13178},
  year={2024}
}

@article{bai2022constitutional,
  title={Constitutional ai: Harmlessness from ai feedback},
  author={Bai, Yuntao and Kadavath, Saurav and Kundu, Sandipan and Askell, Amanda and Kernion, Jackson and Jones, Andy and Chen, Anna and Goldie, Anna and Mirhoseini, Azalia and McKinnon, Cameron and others},
  journal={arXiv preprint arXiv:2212.08073},
  year={2022}
}

@article{jimenez2023swe,
  title={Swe-bench: Can language models resolve real-world github issues?},
  author={Jimenez, Carlos E and Yang, John and Wettig, Alexander and Yao, Shunyu and Pei, Kexin and Press, Ofir and Narasimhan, Karthik},
  journal={arXiv preprint arXiv:2310.06770},
  year={2023}
}

@article{lv2024codeact,
  title={Codeact: Code adaptive compute-efficient tuning framework for code llms},
  author={Lv, Weijie and Xia, Xuan and Huang, Sheng-Jun},
  journal={arXiv preprint arXiv:2408.02193},
  year={2024}
}

@article{bedi2025medhelm,
  title={MedHELM: Holistic Evaluation of Large Language Models for Medical Tasks},
  author={Bedi, Suhana and Cui, Hejie and Fuentes, Miguel and Unell, Alyssa and Wornow, Michael and Banda, Juan M and Kotecha, Nikesh and Keyes, Timothy and Mai, Yifan and Oez, Mert and others},
  journal={arXiv preprint arXiv:2505.23802},
  year={2025}
}

@article{lin2021truthfulqa,
  title={Truthfulqa: Measuring how models mimic human falsehoods},
  author={Lin, Stephanie and Hilton, Jacob and Evans, Owain},
  journal={arXiv preprint arXiv:2109.07958},
  year={2021}
}

@inproceedings{feffer2024red,
  title={Red-teaming for generative AI: Silver bullet or security theater?},
  author={Feffer, Michael and Sinha, Anusha and Deng, Wesley H and Lipton, Zachary C and Heidari, Hoda},
  booktitle={Proceedings of the AAAI/ACM Conference on AI, Ethics, and Society},
  volume={7},
  pages={421--437},
  year={2024}
}

@article{weng2022large,
  title={Large language models are better reasoners with self-verification},
  author={Weng, Yixuan and Zhu, Minjun and Xia, Fei and Li, Bin and He, Shizhu and Liu, Shengping and Sun, Bin and Liu, Kang and Zhao, Jun},
  journal={arXiv preprint arXiv:2212.09561},
  year={2022}
}

@article{wang2022self,
  title={Self-consistency improves chain of thought reasoning in language models},
  author={Wang, Xuezhi and Wei, Jason and Schuurmans, Dale and Le, Quoc and Chi, Ed and Narang, Sharan and Chowdhery, Aakanksha and Zhou, Denny},
  journal={arXiv preprint arXiv:2203.11171},
  year={2022}
}

@article{li2025verifybench,
  title={Verifybench: A systematic benchmark for evaluating reasoning verifiers across domains},
  author={Li, Xuzhao and Li, Xuchen and Hu, Shiyu and Guo, Yongzhen and Zhang, Wentao},
  journal={arXiv preprint arXiv:2507.09884},
  year={2025}
}

@inproceedings{schluter2017limits,
  title={The limits of automatic summarisation according to rouge},
  author={Schluter, Natalie},
  booktitle={Proceedings of the 15th Conference of the European Chapter of the Association for Computational Linguistics},
  pages={41--45},
  year={2017},
  organization={Association for Computational Linguistics}
}

@article{lin2023llm,
  title={Llm-eval: Unified multi-dimensional automatic evaluation for open-domain conversations with large language models},
  author={Lin, Yen-Ting and Chen, Yun-Nung},
  journal={arXiv preprint arXiv:2305.13711},
  year={2023}
}

@inproceedings{papineni2002bleu,
  title={Bleu: a method for automatic evaluation of machine translation},
  author={Papineni, Kishore and Roukos, Salim and Ward, Todd and Zhu, Wei-Jing},
  booktitle={Proceedings of the 40th annual meeting of the Association for Computational Linguistics},
  pages={311--318},
  year={2002}
}

@inproceedings{lin2004rouge,
  title={Rouge: A package for automatic evaluation of summaries},
  author={Lin, Chin-Yew},
  booktitle={Text summarization branches out},
  pages={74--81},
  year={2004}
}

@article{bavaresco2024llms,
  title={Llms instead of human judges? a large scale empirical study across 20 nlp evaluation tasks},
  author={Bavaresco, Anna and Bernardi, Raffaella and Bertolazzi, Leonardo and Elliott, Desmond and Fern{\'a}ndez, Raquel and Gatt, Albert and Ghaleb, Esam and Giulianelli, Mario and Hanna, Michael and Koller, Alexander and others},
  journal={arXiv preprint arXiv:2406.18403},
  year={2024}
}

@article{chen2021finqa,
  title={Finqa: A dataset of numerical reasoning over financial data},
  author={Chen, Zhiyu and Chen, Wenhu and Smiley, Charese and Shah, Sameena and Borova, Iana and Langdon, Dylan and Moussa, Reema and Beane, Matt and Huang, Ting-Hao and Routledge, Bryan and others},
  journal={arXiv preprint arXiv:2109.00122},
  year={2021}
}

@article{jin2021disease,
  title={What disease does this patient have? a large-scale open domain question answering dataset from medical exams},
  author={Jin, Di and Pan, Eileen and Oufattole, Nassim and Weng, Wei-Hung and Fang, Hanyi and Szolovits, Peter},
  journal={Applied Sciences},
  volume={11},
  number={14},
  pages={6421},
  year={2021},
  publisher={MDPI}
}


\end{document}